\documentclass[letterpaper, 10 pt, conference]{ieeeconf}  

\IEEEoverridecommandlockouts                              

\usepackage{times} 
\usepackage{amsmath} 
\usepackage{amssymb}  

\usepackage[subrefformat=parens]{subcaption}
\usepackage{balance}
\usepackage{cite}
\usepackage{graphicx}
\usepackage{booktabs}
\usepackage{tabularx}
\usepackage{multirow}
\usepackage{multicol}
\usepackage{bm}
\usepackage{url}
\usepackage{balance}
\usepackage{color}
\usepackage{bbding}
\usepackage{nccmath}
\usepackage{xspace}
\usepackage{wrapfig}
\usepackage{colortbl}
\usepackage[colorlinks=true,bookmarks=true]{hyperref}
\usepackage[capitalize]{cleveref}
\crefname{section}{Sec.}{Secs.}
\Crefname{section}{Section}{Sections}
\Crefname{table}{Table}{Tables}
\crefname{table}{Tab.}{Tabs.}

\usepackage{comment}
\usepackage{todonotes}
\usepackage{lipsum}

\usepackage{titlesec}
\titleformat{\paragraph}[runin]{\bfseries}{}{1em}{}[:]
\titlespacing*{\paragraph}{0pt}{\baselineskip}{1em}

\usepackage{custom} 
\title{\LARGE \bf Multi-Agent Behavior Retrieval: Retrieval-Augmented Policy Training\\for Cooperative Push Manipulation by Mobile Robots}

\author{So Kuroki$^{\ast,1}$, Mai Nishimura$^{1}$, and Tadashi Kozuno${^1}$
\thanks{$^{\ast}$This work was done while So Kuroki was a research intern at OMRON SINIC X Corporation.}
\thanks{$^{1}$So Kuroki, Mai Nishimura, and Tadashi Kozuno are with OMRON SINIC X Corporation, 5-24-5, Hongo, Bunkyo-ku, Tokyo, Japan
        {\tt\small \{mai.nishimura, tadashi.kozuno\}@sinicx.com}}%
}

\begin{document}

\maketitle
\thispagestyle{empty}
\pagestyle{empty}

\begin{abstract}
Due to the complex interactions between agents, learning multi-agent control policy often requires a prohibited amount of data.
This paper aims to enable multi-agent systems to effectively utilize past memories to adapt to novel collaborative tasks in a data-efficient fashion.
We propose the Multi-Agent Coordination Skill Database, a repository for storing a collection of coordinated behaviors associated with key vectors distinctive to them.
Our Transformer-based skill encoder effectively captures spatio-temporal interactions that contribute to coordination and provides a unique skill representation for each coordinated behavior.
By leveraging only a small number of demonstrations of the target task, the database enables us to train the policy using a dataset augmented with the retrieved demonstrations. Experimental evaluations demonstrate that our method achieves a significantly higher success rate in push manipulation tasks compared with baseline methods like few-shot imitation learning. Furthermore, we validate the effectiveness of our retrieve-and-learn framework in a real environment using a team of wheeled robots.

\end{abstract}

\section{Introduction}
\label{sec:introduction}

\def\DATABASE{MACS-DB}
\looseness=-1
When making decisions or engaging in collaborative tasks such as moving heavy loads~\cite{yang2022collaborative} or playing team sports~\cite{le2017coordinated}, we effectively tap into past experiences by retrieving relevant memories~\cite{mattar2018prioritized}.
By drawing on the wealth of lessons learned from past successes and failures, we gain valuable insights that enhance our coordination ability to work together more efficiently even in unknown situations.
In this paper, we focus on endowing a team of mobile robots with such intelligent memory storage, which we refer to as a \emph{Multi-Agent Coordination Skill Database.}
The database, which serves as a repository of \emph{coordination skills}, efficiently stores collaborative demonstrations in diverse settings. 
Collecting a large number of multi-robot demonstrations for every target configuration is extremely costly and impractical in real-world scenarios.
In contrast, by providing only a few demonstrations of the target task as queries, our database enables us to retrieve and learn from the vast repository of coordination skills that are relevant to the target task.

\begin{figure}[t]
    \begin{center}
    \includegraphics[width=0.95\linewidth]{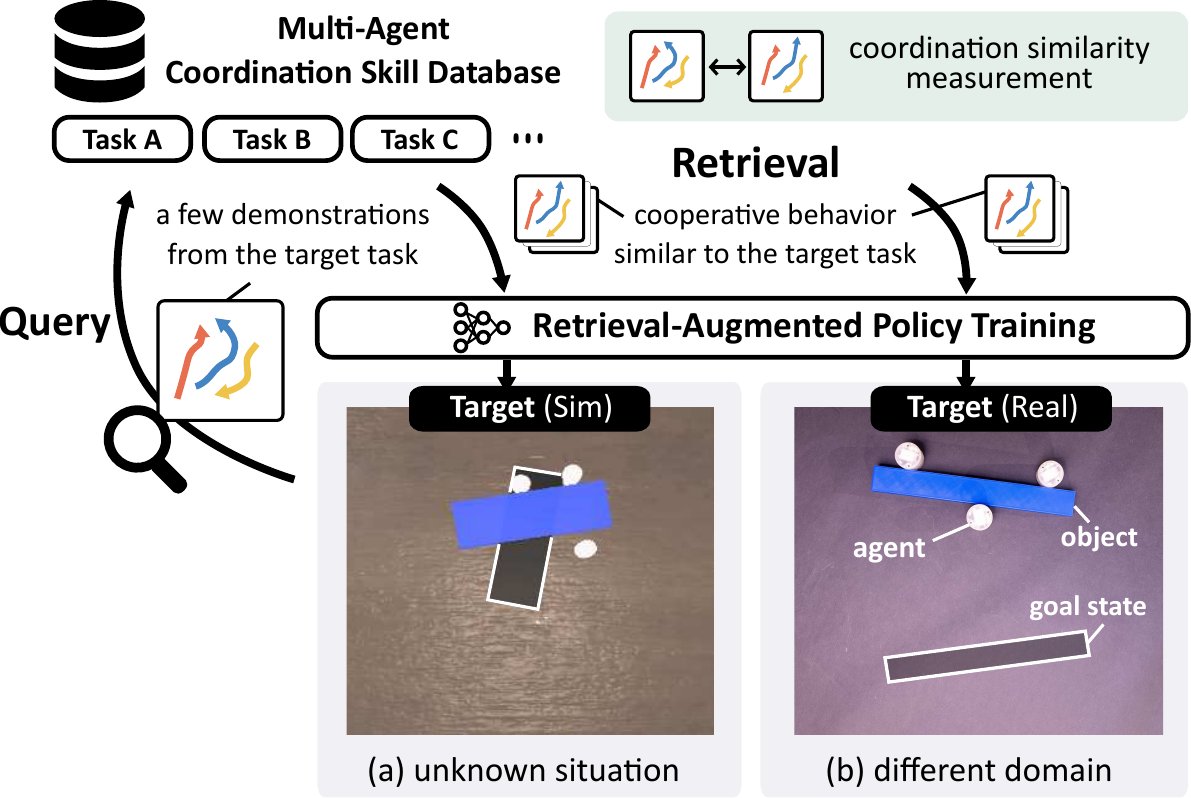}
    \vskip -0.0in
    \caption{Given a collection of unlabeled demonstrations from multiple tasks, our aim is to construct a coordination skill database that enables us to retrieve past experiences for different scenarios or domains (e.g., real-world robots).}
    \label{fig:teaser}
    \end{center}
    \vskip -0.2in
\end{figure}

\looseness=-1
In this study, we explore the potential of such a system in push manipulation by a team of cooperative mobile robots.
While cooperative push manipulation enables more dexterous manipulation and enhances robustness~\cite{shibata2021deep, kuroki2023collective}, capturing complex interactions between agents and an object often requires a prohibited amount of data~\cite{kuroki2023collective} or a specific formulation for each setup~\cite{wang2002object, chen2015occlusion}.
Given the growing interest in cooperative push manipulation in diverse fields such as factory logistics and delivery services, developing data-efficient methods is crucial for its wider application and future advancement.


To achieve data-efficient methods, this study focuses on two research questions:
(i) how to reuse specific coordination behaviors involved in past demonstrations and
(ii) how to efficiently retrieve the specific coordination skills from a large number of unlabeled demonstrations?
While some pioneering works~\cite{nasiriany2023learning,du2023behavior} seek a similar direction of research that retrieves and learns from sub-optimal offline datasets, their studies are limited to single agents with image-based observations.
Unlike single-agent scenarios, multi-agent push manipulations often require both local and global coordination. This involves local cooperation with nearby agents to apply force to an object considering the object's shape, while also making cooperative global decisions on the overall movement direction of the object by multiple robots, such as pushing or rotating.
However, measuring the similarity between two cooperative demonstrations is a non-trivial problem.
The local similarity observed between individual agents does not always conform to the global similarity among all agents.
This discrepancy between local and global similarity poses a challenge for using these demonstrations in downstream policy training.

To address the question, we introduce a novel retrieve-and-learn framework based on the pre-constructed Multi-Agent Coordination Skill Database for push manipulations.
\Cref{fig:teaser} depicts an overview of the proposed concept.
We first train a Transformer-based coordination skill encoder that predicts the next-step actions of agents by considering temporal and spatial interactions among agents and a manipulation object, thereby effectively incorporating local and global interaction dynamics.
The prior demonstrations encompass a variety of scenarios with different object shapes (stick and block) and manipulation directions (pushing and rotating), which are both crucial for robot coordination in push manipulation tasks.
These demonstrations are encoded into a single representative vector and stored in the coordination skill database as a pair of the vector and the original demonstration.
Second, we develop a coordination skill matcher to measure similarities between two cooperative demonstrations in the learned skill representation space. 
Given a small number of target demonstrations, we efficiently retrieve past demonstrations relevant to the target task using the matcher.
Finally, we train the policy for the target task using a dataset augmented with the retrieved demonstrations.
We demonstrate that our approach enables us to retrieve demonstrations that are beneficial for learning the task that requires similar coordination skills in both simulated and real-world scenarios.

The contributions of this paper are summarized as follows.
\begin{itemize}
\item We introduce a novel retrieve-and-learn framework that enables us to train a multi-agent control policy using only a small number of target demonstrations. 
\item We introduce the Multi-Agent Coordination Skill Database, a vector-based database for storing and retrieving push manipulation behaviors, covering different object shapes and manipulation directions associated with a specific feature vector.
\item We demonstrate that the skill database can provide reusable skill representations not only for a simulated push manipulation task but also for a team of real-wheeled robots.
\end{itemize}


\section{Related Work}
\label{sec:related_work}

\subsection{Multi-Agent Push Manipulations}
Multi-agent push manipulation has received significant interest due to its substantial potential in object transportation, particularly in factory logistics and delivery services.
In this task, it is crucial to acquire a complex interaction model that captures the relationships between agents and an object.
Earlier research primarily concentrated on explicitly modeling interactions between robots and objects~\cite{wang2002object} or using rule-based methods to define robot movements~\cite{chen2015occlusion}. 
More recent studies have shifted
towards deep learning-based approaches, which leverage reinforcement learning or imitation learning to obtain cooperative interaction models~\cite{shibata2021deep, kuroki2023collective, nachum2019multi, wu2021spatial}. 
While these methods work well in simulated scenarios,
they generally necessitate large datasets or long-time explorations, posing a challenge for real-world applications. In response, we introduce the Multi-Agent Coordination Skill Database based on interaction models distinctive to task-specific coordinated behaviors.
The database enables the training of a multi-agent control policy using only a limited set of target demonstrations even in real-world situations.

\subsection{Imitation Learning from Prior Demonstrations}
Learning policies from experts have been extensively studied in various domains such as vehicle control~\cite{bhattacharyya2018multi} and multi-agent navigation~\cite{renz2022plant}.
Two major drawbacks of these approaches are that the policy suffers from distribution shifts when adapted to unknown situations, and that the policy performs poorly when trained on sub-optimal demonstrations.
Few-shot imitation learning is a promising approach for fine-tuning a policy, initially trained with a large-scale, multi-task dataset, to adapt to the target task using a small number of demonstrations~\cite{mandi2022towards, fang2023rh20t}. 
In contrast, the retrieval approach introduces a novel direction by using optimal demonstrations as a query to retrieve demonstrations relevant to the target task~\cite{du2023behavior, nasiriany2023learning}.
Our work is built on top of their retrieval concept and is the first to expand this retrieve-and-learn paradigm to a multi-agent behavior database.

\subsection{Constructing Multi-Task Prior Database}
Following the success of large-scale pre-trained models built on extensive datasets, the robot learning community has developed open-source datasets~\cite{mandlekar2018roboturk, fu2020d4rl, open_x_embodiment_rt_x_2023}.
These datasets feature demonstrations spanning multiple tasks and a range of robot types, gathered through methods like teleoperation~\cite{mandlekar2018roboturk}, diverse learning approaches~\cite{fu2020d4rl}, and a combination of these methods inclusive of optimization techniques~\cite{open_x_embodiment_rt_x_2023}. 
Despite these advancements, similar dataset development is not as prevalent in the multi-agent domain, particularly due to the inherent difficulty in controlling multi-agent robots associated with collision avoidance~\cite{huang2019collision}.
To address these challenges, our work uses reinforcement learning in conjunction with the hybrid policy design incorporating reciprocal velocity obstacles (RVO)~\cite{van2008reciprocal} to collect collision-free demonstrations of multiple agents at scale.


\begin{figure*}[t]
    \begin{center}
    \includegraphics[width=0.95\linewidth]{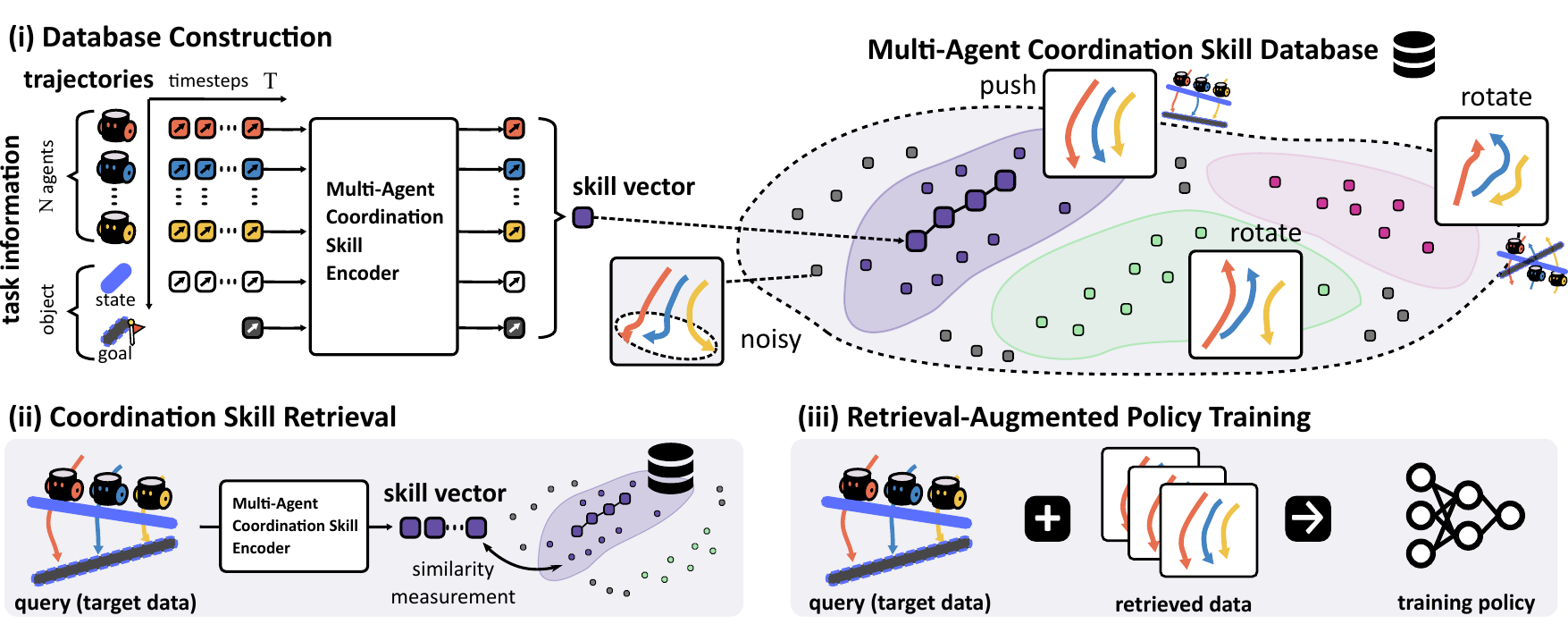}
    \vskip -0.0in
    \caption{Our retrieve-and-learn framework consists of
three primary components: (a) constructing the Multi-Agent Coordination Skill
Database on the basis of the prior experiences, (b) retrieving
demonstrations using a few target demonstrations as queries, and (c) learning the multi-agent control policy using the retrieved data and target data.}
    \label{fig:method}
    \end{center}
    \vskip -0.2in
\end{figure*}

\section{Problem Formulation}
\label{sec:formulation}
\looseness=-1
We consider a team of mobile agents working on a collaborative task, \eg pushing an object toward its goal while interacting with each other. Given a large task-agnostic prior dataset \DATASET{prior} and a few demonstrations \DATASET{target} collected from the target task, our main objective is to retrieve coordination skills from \DATASET{prior} that facilitate downstream policy learning for the target task.

\subsection{Multi-Agent Coordination Skill Database}

\DATASET{prior} consists of task-agnostic demonstrations of $N$ mobile agents.
Each demonstration, $\traj{i} = \left ( \mathcal{O}^1, \dots, \mathcal O^{T_i}\right )$, is a sequence of variable length $T_i$,
where $\mathcal O^t = \{\left (\bm s_n^t, \bm a_n^t\right )\}_{n=1}^N$ is a set of $N$ agents' state-action pairs for time step $t$.
To effectively retrieve demonstrations from \DATASET{prior} that are relevant to \DATASET{target}, we seek an abstract representation to measure the similarity between the two multi-agent demonstrations.
We refer to this as \emph{multi-agent coordination skill representation}, a compressed vector representation that is distinctive to the specific coordination behavior.
For N agents' states $\bm s^t$, we aim to learn a skill encoder $\mathcal E$ that maps the state representation of multiple agents into a single representative vector $\skill \in \mathbb R^n$.
Once the coordination skill representation space is learned on the prior dataset \DATASET{prior}, we construct a Multi-Agent Coordination Skill Database, a vector database consisting of the skill representation $\skill$ associated with the original state-action pairs $\mathcal O$.

\subsection{Retrieval-Augmented Policy Training}
We assume that the prior dataset \DATASET{prior} and the target demonstrations \DATASET{target} are composed as follows,
\begin{itemize}
    \item{\DATASET{prior}} : 
    The prior dataset encompasses a large-scale offline demonstration of diverse cooperative tasks. 
    The dataset also includes noisy or sub-optimal demonstrations to mimic real-world scenarios. All demonstrations are task-agnostic, \ie each data is stored without any specific task annotations.
    \item{\DATASET{target}}: The target dataset encompasses a small amount of expert data, \ie all data is composed of well-coordinated demonstrations to complete the target task.
\end{itemize}

Given \DATASET{prior} and \DATASET{target}, we aim to retrieve cooperative behaviors similar to those seen in \DATASET{target} in the learned coordination skill space.
Once we obtain the retrieved data \DATASET{ret}, we train a multi-agent control policy using \DATASET{target} augmented with \DATASET{ret}. That is, the training data \DATASET{train} is described as \DATASET{train} $=$ \DATASET{target} $\cup$ \DATASET{ret}.

\subsection{Task Formulation}
We consider the task of cooperative push manipulation where the objective is to navigate $N$ mobile agents to push an object from the start to the goal while also interacting with each other.
We assume a centralized policy~\cite{renz2022plant} for controlling these agents and use imitation learning to train the policy. The objective of imitation learning is to learn a policy $\pi$ that imitates the cooperative behavior of an expert $\pi^*$.
Given offline demonstrations composed of state-action pairs $\mathcal O$, we aim to learn a cooperative policy $\pi: \bm s^t \mapsto \bm a^{t}$ that maps the multi-agent states to a set of their future actions.
The state $\state{t}{}$ is defined as $\state{t}{} = \left [ \{\bm r^t_n\}_{n=1}^N, \bm o^t, \bm g\right]$, where $\bm r^t_n$ denotes each agent's state and $\bm o^t$ is that of a manipulation object, respectively. We set the goal state $\bm g$ as a constant in a sequence. 
Each agent state $\bm{r}$ is defined as
\begin{equation}
\bm{r} = [\bm{p}, \bm{q_{rot}}, w, h, \bm{v}_{\mathrm{prev}}]\,,
\end{equation}
where \( \bm{p} \) is the 2D position of the agent relative to the goal state coordinate, \( \bm{q_{rot}} \) is rotation to the world coordinate system represented by quaternion,
\( w \) and \( h \) specify the width and height of the bounding box defined in the xy-plane in accordance with an agent, and \( \bm{v}_{\text{prev}} \) is 2D velocity in the previous timestep.
The object state $\bm o$ follows the same state components as the agent $\bm r$.
The next-step action $\bm a^t$ of each agent is defined as 2D velocity.

\begin{figure}[t]
    \begin{center}
    \includegraphics[width=\linewidth]{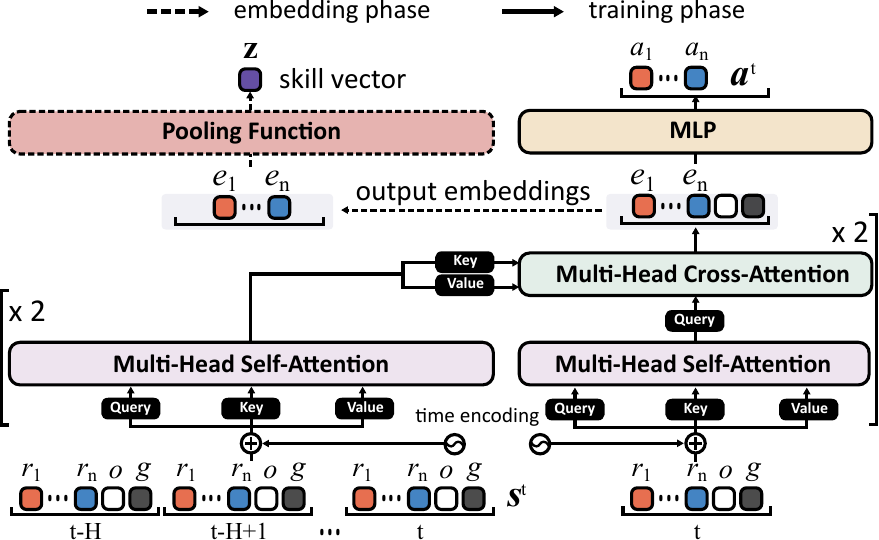}
    \vskip -0.0in
    \caption{
    In the training phase, our model takes past trajectory sequences as input tokens and outputs the future trajectories \ie actions of each input token. In the embedding phase, the model pools the multi-agent feature embeddings into a single representative vector.}
    \label{fig:model}
    \end{center}
    \vskip -0.2in
\end{figure}

\section{Proposed Method}
\label{sec:proposed_method}
\Cref{fig:method} illustrates an overview of our proposed framework. 
The primary challenge of our retrieve-and-learn framework is to develop coordination skill representation that can discern similar or dissimilar coordinated behaviors of multiple agents.
To address this, we propose a coordination skill encoder based on an attention mechanism.
The encoder learns interactions between agents and a manipulation object from diverse prior demonstrations.

Furthermore, robot collisions, especially in the real world, pose a significant challenge in tasks that require multi-agent coordination. We propose a controller that integrates a learning-based high-level policy with an optimization-based low-level policy. 
This design ensures safety by effectively avoiding robot collisions and makes our overall framework applicable to diverse environments.

\subsection{Multi-Agent Coordination Skill Encoder}

We introduce a Transformer-based coordination skill encoder, which learns to capture interactions among agents as well as interactions between agents and a manipulation object.
The design of the coordination skill encoder is guided by the following key insights.
First, the coordination occurs in both spatial and temporal domains.
For example, if one agent maneuvers to the right side to contact an object, the other one must contact the left side of the object while predicting and interpreting the intention of the first agent.
Second, the skill representation not only encompasses past trajectories but also predicts future trajectories. This enables the coordination skill to guide downstream policies, suggesting the appropriate action to take in future timesteps.
To obtain such predictable skill representation space, we train the Transformer-based skill encoder as a prediction model of the agent's future trajectories \ie actions.

\vspace{-12pt}
\paragraph{Model Overview}
\Cref{fig:model} illustrates an overview of our proposed model.
We follow an encoder-decoder design of the standard Transformer architecture~\cite{vaswani2017attention}, where the encoder summarizes contextual information, while the decoder predicts the next state of the input tokens, taking into account the encoded context.
In our network, the past context encoder is comprised of stacked multi-head self-attention layers that learn to attend to past trajectories across spatial and temporal domains. The future trajectory decoder is composed of multi-head self-attention layers and subsequent stacked multi-head cross-attention layers that integrate the past trajectory information and input tokens.
Given past trajectory sequences with a temporal time window $H$, 
$\mathbf X = \left ( \bm s^{t-H}, \bm s^{t-H+1}, \dots, \bm s^t \right )$, 
our model is trained to predict a set of next-step actions of the agents $\mathbf Y = \{\bm a^{t}_n\}_{n=1}^N$.
\vspace{-12pt}
\paragraph{Attention Mechanism for Spatio-Temporal Interaction Encoding}
As Transformers take sequential data as inputs, we first decompose trajectories of multiple agents, the manipulation object, and the goal state into individual tokens across agents, an object, a goal state, and time domains.
The attention mechanism takes three types of inputs: a set of queries $\bm Q$, keys $\bm K$, and values $\bm V$.
Formally, the attention function can be written as the mapping of a query and a set of key-value pairs to an output,
\begin{equation}
\mathrm{Attn}(\bm Q,\bm K,\bm V) = \mathrm{softmax}\left (\frac{\bm Q^\top \bm K}{\sqrt {d_k}}\right )\bm V\,,
\end{equation}
where $\sqrt d_k$ is a scaling factor set in accordance with the length of the tokens.
Each query, key, and value vector $\bm q = W_q \bm h_s$, $\bm k = W_k \bm h_t$, $\bm v = W_v \bm h_t$ are calculated from the input state sequence $\bm h$ by a linear projection matrix $W$.
We refer to the case of $\bm h_s = \bm h_t$ as self-attention and $\bm h_s \neq \bm h_t$ as cross-attention.
In principle, the attention function is permutation invariant to the input tokens.
To inform the model of the time sequence order, we add time encoding to the input embeddings by sinusoidal encoding.
In our setting, since the input tokens include positional information, the additional time encoding enables us to compute attention over spatial and temporal domains.

\vspace{-12pt}
\paragraph{Training Loss}
In the training phase, the obtained set of features $\bm E = \lbrace \bm e_1, \ldots, \bm e_N \rbrace $ is decoded to a set of future actions by the multi-layer perception (MLP) layer.
The skill encoder is trained to minimize the mean squared error (MSE) between ground truth actions $a$ and predicted actions $\hat{a}$:
\begin{equation}
    \text{MSE} = \frac{1}{T_i} \sum_{t=1}^{T_i}  (a^t - \hat{a}^t)^2.
\end{equation}
We train the coordination skill encoder using the whole \DATASET{prior} involving diverse interactions by varying the number of agents.
Our Transformer-based model accepts input tokens with variable lengths and is permutation-invariant in principle, which enables us to capture diverse interactions that are not limited to a specific number of agents.

\vspace{-12pt}
\paragraph{Multi-Agent Feature Pooling}
In the embedding phase, we pack the obtained features $\bm E = \lbrace \bm e_1, \ldots, \bm e_N \rbrace$ into a single representative vector $\skill$, \ie skill representation.
To extract a representative feature invariant to both the specific number and order of agents, we employ a symmetric function~\cite{qi2017pointnet}. The function $\phi(\cdot)$ takes a set of vectors as input and outputs a single vector invariant to the input order,
\begin{equation}
\skill = \phi \left (\{\bm e_1, \ldots , \bm e_N \} \right)\,.
\end{equation}
For the symmetric function $\phi$, we use an average pooling operation since it demonstrated better separation performance in the learned representation space compared with the max pooling operation.
Finally, the Multi-Agent Coordination Skill Database is constructed by a collection of the skill vector $\bm z$ associated with the original state-action pairs $\mathcal O$.

\subsection{Data Retrieval from the Multi-Agent Coordination Skill Database}
Given the query sampled from \DATASET{target}, our goal is to retrieve demonstrations that exhibit similar coordinated behavior to the query from the coordination skill database constructed from \DATASET{prior}.
We first encode the query demonstrations into the learned skill representation space by the pre-trained skill encoder $\mathcal E$, and compare the similarity between the embedded query and collection of skill vectors in the database.
The database stores sequences of skill vectors \( \mathcal{Z}_j = \left( \skill^1, \dots, \skill^{T_i} \right) \) with a variable length \(T_i\) embedded from diverse demonstrations in \DATASET{prior}.
By encoding the query demonstrations through the skill encoder $\mathcal E$, we obtain a sequence of skill vectors with a variable length \(T'_j\), \( \mathcal{Z}'_j = \left( \skill'^1, \dots, \skill'^{T'_j} \right) \).
Note that these two sequences, \( \mathcal{Z}_i \) and \( \mathcal{Z}'_j \), may have different lengths \( T_i \) and \( T'_j \). 
To compare the two sequences of disparate lengths, we use dynamic time warping (DTW)~\cite{muller2007dynamic}, a renowned technique that can accommodate sequences of differing lengths to calculate the similarity between temporal sequences.
\begin{equation}
DTW(\mathcal{Z}_i, \mathcal{Z}'_j) = \min_{\alpha \in A(\mathcal{Z}_i, \mathcal{Z}'_j)} \left( \sum_{(i,j)\in \alpha} m(\skill^i, \skill'^j) \right)\,,
\end{equation}
where $m$ is the distance metric between the two elements, $A$ contains all potential alignment paths between the two sequences, and $\alpha$ is the alignment path that pairs elements between sequences.
In the experiment, we adopt a particular variant of DTW named \textit{FastDTW}~\cite{salvador2007toward}, which is recognized for its ability to approximate DTW in linear time, making it scalable for larger datasets.
Assuming \DATASET{prior} and \DATASET{target} have \( L^p \) and \( L^t \) number of demonstrations, respectively, we compute the similarity for all combinations.
\begin{align}
\textit{FastDTW}(\mathcal{Z}_i, \mathcal{Z}'_j) \, \forall \, i &\in \{0, 1, \dots, L^p-1\} \nonumber \\
\text{and} \, j &\in \{0, 1, \dots, L^t-1\}
\end{align}
We utilize the cosine similarity as the distance metric $m$.
We identify the top $K$ similar demonstrations from \DATASET{prior} to each target demonstration in \DATASET{target}, thereby retrieving a total of $L^t \times K$ demonstrations as \DATASET{ret} from \DATASET{prior}.

\subsection{Hierarchical Policy for Safe Robot Control}

Our ultimate goal is to deploy the cooperative policy $\pi$ trained on the simulated dataset to real-world robots~\cite{le2016zooids}.
We employ a hierarchical policy design to ensure that our trained policy can navigate real-world mobile robots safely.
The hierarchical planning policy is composed of i) a high-level policy that generates the agent's velocities at sparse intervals and ii) a low-level control policy that adjusts the velocities outputted from the high-level policy to avoid potential collisions.
We set the high-level control policy as our trained policy $\pi$ and the low-level control policy as an off-the-shelf collision avoidance algorithm.
We use the RVO algorithm with nonholonomic constraint~\cite{snape2010smooth}.
That is, our policy can focus on the high-level planning of the cooperative behavior of each agent and can safely delegate collision avoidance to the external algorithm.
This design enables us to make our overall framework versatile for typical wheeled robots with nonholonomic kinematic constraints.

\section{Experiment}
\label{sec:experiment}

In this section, we experimentally evaluate our method using both simulation and real-world scenarios. 
Through our experiments, we aim to address the following questions:
\begin{itemize} 
    \item How is our multi-agent representation distinctly characterized for each skill? (\ref{subsec:representation})
    \item Can our method retrieve coordinated behaviors similar to those in the prior dataset? (\ref{subsec:retrieve})
    \item How does the retrieved dataset contribute to improving the success rate of the target task? (\ref{subsec:policy} and \ref{subsec:ablation})
    \item Can our method also enhance the control of multi-agent robots in real-world settings? (\ref{subsec:real})
\end{itemize}

\subsection{Tasks and Dataset}
\label{subsec:setup}
\paragraph{Task Design}
We assume $N$ mobile robots are navigated to push an object toward a predefined goal state.
To explore various coordination scenarios, we specifically focus on three different values of $N \in \{2,3,4\}$.
For each of these setups, we collect demonstrations that involve four distinct tasks.
These four tasks are carefully designed to encompass two different objects (\textbf{stick} or \textbf{block}) and manipulation difficulties (\textbf{easy} or \textbf{hard}).
The stick and block objects are designed as 2D rectangles with dimensions of $0.24\,\mathrm{m}$ by $0.02\,\mathrm{m}$ and $0.24 \,\mathrm{m}$ by $0.08\,\mathrm{m}$, respectively.
Stick objects can only be pushed stably from their longer side, while block objects can be pushed from either their longer or shorter side. These two different object designs result in diverse coordination scenarios.
The difficulty of the task is defined by the relative distance $d$ and the relative rotation $\theta$ between the initial and goal states.
We refer to a \textbf{simple object rotating} task as \textbf{easy}, where the goal state is close to the initial state, ranging from $0$ to $0.1\,\mathrm{m}$, and $\theta$ ranges from $-90^\circ$ to $90^\circ$.
A \textbf{combination of pushing and rotating} tasks is referred to as \textbf{hard}. As for hard tasks, the goal state is placed further from the initial state, with distances ranging from $0.1$ to less than $0.2\,\mathrm{m}$ and $\theta$ ranges from $-90^\circ$ to $90^\circ$.
At every setup, the initial position of the object is fixed, while the initial positions of the robots and the goal state are randomly sampled.

\textbf{Dataset:}
We collect both the prior data of \DATASET{prior} and target data of \DATASET{target} in the simulated environment implemented by Unity ML Agents~\cite{juliani2020}.
We develop reinforcement learning-based policies based on soft actor-critic (SAC)~\cite{haarnoja2018soft} trained at various proficiency levels.
To mimic real-world datasets involving sub-optimal demonstrations collected by non-experts, we prepared three distinct policies—classified as best, average, and worst—on the basis of their reward. 
This classification enabled the collection of a diverse dataset that reflects a spectrum of policy effectiveness. 
As for \DATASET{prior}, we amassed $2,500$ demonstrations from each policy. 
For each $N$, this equates to $2,500 \times 3$ policies $\times 4$ tasks, totaling $30,000$ demonstrations.
As for \DATASET{target}, $250$ demonstrations were collected from the best policy for each task.

\subsection{Baselines}
To the best of our knowledge, our work is the first to propose retrieving and learning coordination skill representations for multi-agent systems.
Due to the lack of directly comparable existing methods, we have established two alternative baselines to assess our proposed method.
\vspace{-12pt}
\paragraph{Agent-wise trajectory matching (A-TM)}
We utilize trajectory matching as a baseline to evaluate our retrieval method on the basis of multi-agent skill representation. 
By calculating the similarity of the robot's xy-coordinate trajectories using \textit{FastDTW}, we retrieve data from the \DATASET{prior} that have trajectories similar to the target data's trajectories.\vspace{-12pt}
\paragraph{Few-shot Imitation Learning (F-IL)~\cite{du2023behavior, nasiriany2023learning}}
We compare our method with the adaptation method.
We trained the multi-task policy from \DATASET{prior} and fine-tuned it using \DATASET{target}.

\subsection{Training Details}
\label{subsec:parameters}
In the implementation of our coordination skill encoder, we use two self-attention layers in the past context encoder and also two self-attention and cross-attention layers in the future trajectory decoder.
We set the number of heads to $8$ in all multi-head attention layers.
The embedded dimensions of queries $\bm Q$, keys $\bm K$, and values $\bm V$ are all set to $256$.
We use an MLP layer with hidden dimensions of $512$ and $256$ to output the future trajectories.
In the training phase, we use the Adam optimizer~\cite{kingma2014adam}. The learning rate and dropout rate are set to $1e-4$ and $0.1$, respectively. 
We utilize the same model architecture for both the skill encoder and the multi-agent control policy.
We trained the skill encoder with $10$ epochs and all control policies with $50$ epochs.

\subsection{Visualization of the Multi-Agent Coordination Skill Representation Space}
\label{subsec:representation}

\begin{figure*}[t]
    \begin{center}
    \includegraphics[width=0.85\linewidth]{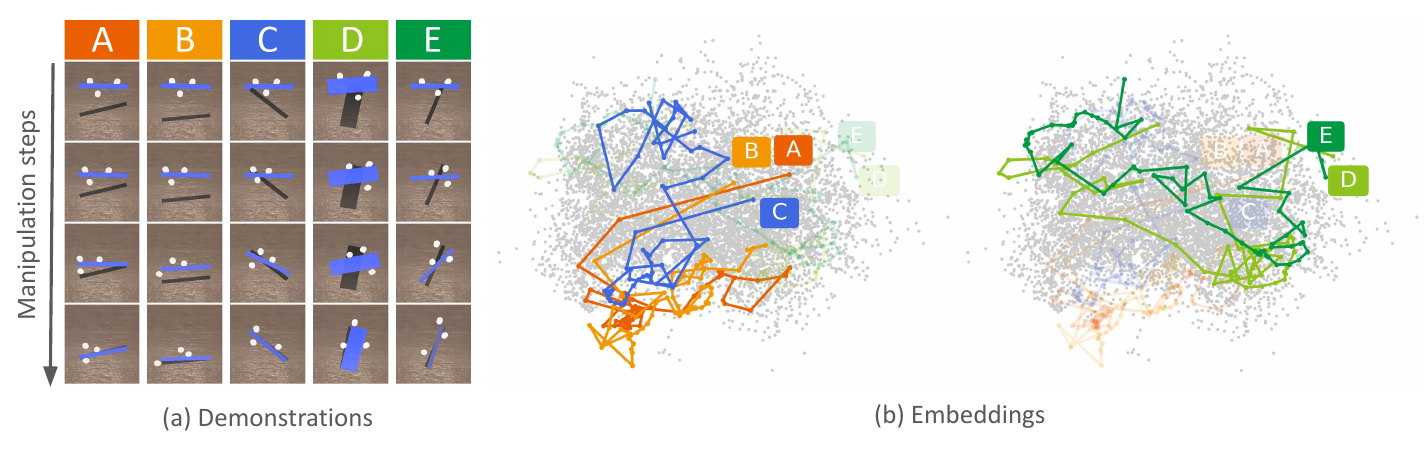}
    \vskip -0.1in
    \caption{(a) Visualization of five different trajectories (N=3): A, B, C, and E represent tasks of pushing a stick at a hard level, while D represents the task of pushing a block at a hard level. (b) Visualization of Multi-Agent Coordination representation. The left image highlights the representations of A, B, and C, while the right image highlights those of D and E. Each trajectory and its corresponding representation sequence are matched with both a unique color and an alphabet ID. 
    }
    \label{fig:representation}
    \end{center}
    \vskip -0.2in
\end{figure*}
To accurately retrieve demonstrations that are similar to the target query, the skill representation should exhibit similarity in coordination behavior, while being dissimilar for unrelated tasks based on the object shapes and manipulation directions.
We project each coordination skill representation into lower-dimensional space to view a property of the learned representation space.
\Cref{fig:representation} visualizes the first two dimensions of the embedding space projected by principal component analysis (PCA).
\Cref{fig:representation} (a) shows five example sequences of demonstrations, and \Cref{fig:representation} (b) highlights their corresponding embeddings.
\vspace{-12pt}
\paragraph{Similar Coordination Behaviors}
In the left image of (b), A and B have the same representation space. This is because their behaviors are similar, as their goal states are alike. Observing the representations of A and C, they initially appear similar but later diverge. 
Trajectories A and C share similarities in coordination until the midpoint; beyond this, they diverge: A starts rotating the object counterclockwise, while C starts rotating it clockwise. Consequently, the representations of A and C initially mirror each other but later differ, reflecting their distinct behaviors.
\vspace{-12pt}
\paragraph{Dissimilar Coordination Behaviors}
Comparing the left and right images, A, B, and C occupy distinctly separate spaces from D and E, with minimal overlap. This reflects the difference in trajectories between A, B, C and D, E.
\vspace{-12pt}
\paragraph{Different Tasks}
In the right images of (b), the representations of D and E have substantial overlap. Although D and E are pushing different objects, a stick and a block, respectively, the similarity in coordination is reflected in their overlapped representations.

\subsection{Qualitative Results of Retrieved Data}
\label{subsec:retrieve}
\begin{figure}[t]
    \begin{center}
    \includegraphics[width=0.95\linewidth]{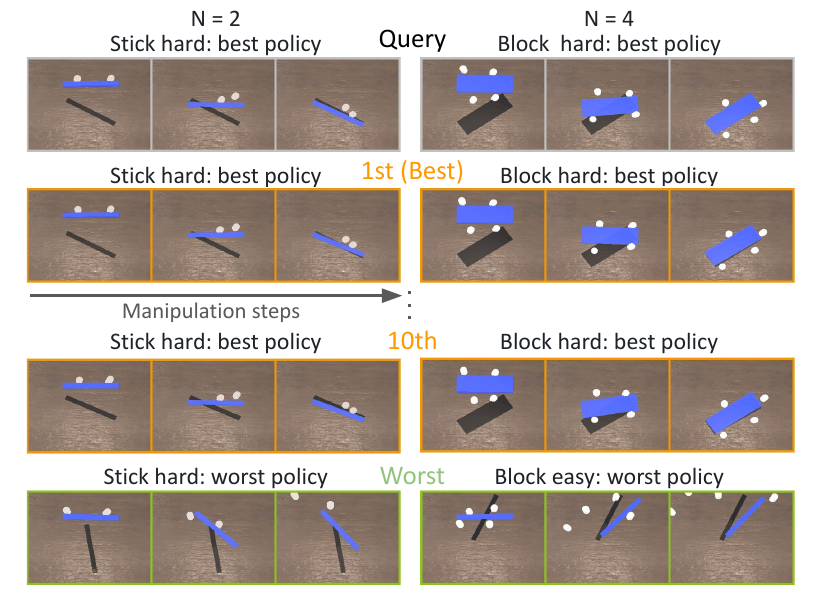}
    \vskip -0.0in
    \caption{The retrieved data is visualized with the two queries sampled from \DATASET{target}. 
    For each demonstration, we display the manipulated object, the task difficulty, and the policy that generated the trajectory.}
    \label{fig:retrieve}
    \end{center}
    \vskip -0.2in
\end{figure}
In our retrieve-and-learn framework, all the retrieved demonstrations are ranked with the similarity metric, and the top-$10$ ranked demonstrations are used for downstream policy training.
\Cref{fig:retrieve} shows demonstrations ranked from the $1$\textsubscript{st} (best) to $10$\textsubscript{th} and worst retrieved by the query demonstration sampled from \DATASET{target}.
The retrieved demonstrations ranked from $1$\textsubscript{st} to $10$\textsubscript{th} show similar movements to the query data. 
The demonstrations with the worst similarity possess goal states differing from the query data. 
Additionally, their trajectories are generated by the worst policy, which lacks effective coordination to complete tasks successfully.
The result demonstrates that our method is capable of correctly ranking related movements higher and unrelated movements lower in relevance to the query data.

\subsection{Evaluation of Downstream Policy Learning}
\label{subsec:policy}
\Cref{tab:baseline} shows our retrieval-augmented policy training greatly outperforms F-IL~\cite{du2023behavior, nasiriany2023learning}.
F-IL's lower accuracy can likely be attributed to the multi-task policy, being adversely influenced by the noise data in \DATASET{prior}, and to the method's inability to sufficiently fine-tune for the distribution shift with the limited data of \DATASET{target}.
Our method also outperforms A-TM. As the number of robots $N$ increases, the performance gap between our method and A-TM widens. 
This indicates that our approach is more effective in tasks requiring advanced robot coordination, particularly in scenarios with a larger number of robots or in more complex tasks.

\begin{center}
\begin{table}
  \vspace{6pt}
 \centering
  \caption{\textbf{Comparison with Baselines.} We evaluate the policies with the 70 new test environments and calculate the success rate and standard error of the mean. The higher success rate and lower standard error of the mean are better.}
  \label{tab:baseline}
  \begin{tabularx}{0.83\linewidth}{cccccc}
    \toprule
    &\multicolumn{2}{c}{\textbf{Task}} & \textbf{Ours} & \textbf{A-TM} & \textbf{F-IL} \\
    N & object & level & (\%) & (\%) & (\%) \\
    \midrule
    2 &block& hard   & \textbf{41.4$\pm$5.9} & \textbf{41.4$\pm$5.9} & 20.0$\pm$4.8 \\
      &block& easy   & \textbf{57.1$\pm$6.0} & 55.7$\pm$6.0 & 15.7$\pm$4.4 \\
      &stick& hard   & \textbf{35.7$\pm$5.8} & 32.9$\pm$5.7 &  5.7$\pm$2.8 \\
      &stick& easy   & 67.1$\pm$5.7 & \textbf{68.6$\pm$5.6} & 12.9$\pm$4.0 \\
    \midrule
    3 &block& hard   & 42.9$\pm$6.0 & \textbf{48.6$\pm$6.0} & 32.9$\pm$5.7 \\
     &block &easy   & \textbf{90.0$\pm$3.6} & 77.1$\pm$5.1 & 34.3$\pm$5.7 \\
     &stick& hard   & \textbf{41.4$\pm$5.9} & 28.6$\pm$5.4 & 34.3$\pm$5.7 \\
     &stick &easy   & 65.7$\pm$5.7 & \textbf{67.1$\pm$5.7} & 52.9$\pm$6.0 \\
    \midrule
    4 &block& hard   & \textbf{58.6$\pm$5.9} & 55.7$\pm$6.0 & 38.6$\pm$5.9 \\
     &block& easy   & \textbf{88.6$\pm$3.8} & 78.6$\pm$4.9 & 55.7$\pm$6.0 \\
     &stick& hard   & \textbf{24.3$\pm$5.2} & 18.6$\pm$4.7 &  7.1$\pm$3.1 \\
     &stick& easy   & \textbf{70.0$\pm$5.5} & 52.9$\pm$6.0 & 57.1$\pm$6.0 \\
    \midrule
    \midrule
    \rowcolor[HTML]{E6E6EF}
    \multicolumn{3}{c}{\textbf{ALL}} & \textbf{56.9$\pm$1.7} & 52.1$\pm$1.7 & 30.6$\pm$1.6 \\
    \bottomrule
  \end{tabularx}
  \vskip -0.2in
\end{table}
\end{center}

\vspace{-10mm}
\subsection{Ablation Study}
\label{subsec:ablation}
\paragraph{Efficacy of Retrieval-Augmented Policy Training}
We conduct an ablation study to evaluate the effectiveness of our methods.
To assess the role of the retrieval method, we train the policies using data solely from \DATASET{target} (Target) and \DATASET{target} $\cup$ \DATASET{prior} (ALL).
\Cref{tab:ablation} shows that our method significantly outperforms these approaches.
This suggests that our method, by appropriately retrieving data from \DATASET{prior} on the basis of \DATASET{target}, is able to augment the training strategy effectively.
\vspace{-12pt}
\paragraph{Distance Metric}
We use the Euclidean distance as a similarity metric for searching and training the policy with the retrieved data. 
While the performance is fifty-fifty for each task, an overall comparison reveals that the proposed method, which employs cosine distance as the similarity metric, outperforms the policy using the Euclidean distance.
\vspace{-22pt}
\paragraph{Balancing Loss Weights}
Moreover, we modified the loss weights for behavior cloning on the basis of the similarity of the retrieved data. 
Given a list of similarity of the retrieved data ${l_1, l_2, ..., l_K}$, the scaled loss weight is calculated as $S_i = \frac{l_{\text{min}}}{l_i}$ for $i = 1, 2, ..., K$, where $S_i$ is the scaled loss weight corresponding to $l_i$.
The result shows that the performance of this weighted behavior cloning is declining. 
This may be because the weighting has led to an undervaluation of beneficial trajectories during training.

\begin{table}
 \centering
  \caption{\textbf{Ablation study}. We evaluate the policies with the 70 new test environments and calculate the success rate and standard error of the mean. The higher mean and lower standard error of the mean are better.}
  \label{tab:ablation}
  \setlength{\tabcolsep}{0.65mm}
  \begin{tabularx}{0.95\linewidth}{cccccccc}
    \toprule
    &\multicolumn{2}{c}{\textbf{Task}} & \textbf{Ours} & \textbf{ALL} & \textbf{Target} & \textbf{Euclidean} & \textbf{Weight} \\
    N & object & level & (\%) & (\%) & (\%) & (\%) & (\%) \\
    \midrule
    2 &block& hard   & \textbf{41.4$\pm$5.9} & 15.7$\pm$4.4 & 11.4$\pm$3.8 & 35.7$\pm$5.8          & 35.7 $\pm$ 5.8 \\
      &block& easy   & \textbf{57.1$\pm$6.0} & 22.9$\pm$5.1 & 17.1$\pm$4.5 & 47.1$\pm$6.0          & 44.3 $\pm$ 6.0 \\
      &stick& hard   & 35.7$\pm$5.8          & 17.1$\pm$4.5 & 18.6$\pm$4.7 & \textbf{38.6$\pm$5.9} & 34.3 $\pm$ 5.7 \\
      &stick& easy   & 67.1$\pm$5.7          & 32.9$\pm$5.7 &  7.1$\pm$3.1 & \textbf{70.0$\pm$5.5} & 65.7 $\pm$ 5.7 \\
    \midrule
    3 &block& hard   & 42.9$\pm$6.0          & 14.3$\pm$4.2 & 12.9$\pm$4.0 & \textbf{57.1$\pm$6.0} & 45.7 $\pm$ 6.0 \\
     &block &easy    & \textbf{90.0$\pm$3.6} & 24.3$\pm$5.2 & 11.4$\pm$3.8 & 81.4$\pm$4.7          & 81.4 $\pm$ 4.7 \\
     &stick& hard    & 41.4$\pm$5.9          & 21.4$\pm$4.9 &  2.0$\pm$2.0 & \textbf{55.7$\pm$6.0} & 27.1 $\pm$ 5.4 \\
     &stick &easy    & \textbf{65.7$\pm$5.7} & 45.7$\pm$6.0 & 20.0$\pm$4.8 & 51.4$\pm$6.0          & 64.3 $\pm$ 5.8 \\
    \midrule
    4 &block& hard   & 58.6$\pm$5.9          &  8.6$\pm$3.4 & 12.9$\pm$4.0 & \textbf{60.0$\pm$5.9} & 48.6 $\pm$ 6.0 \\
     &block& easy    & \textbf{88.6$\pm$3.8} & 51.4$\pm$6.0 & 21.4$\pm$4.9 & 85.7$\pm$4.2          & 77.1 $\pm$ 5.1 \\
     &stick& hard    & 24.3$\pm$5.2          &  1.4$\pm$1.4 &  4.3$\pm$2.4 & \textbf{25.7$\pm$5.3} & 18.6 $\pm$ 4.7 \\
     &stick& easy    & 70.0$\pm$5.5          & 25.7$\pm$5.3 &  8.6$\pm$3.4 & 55.7$\pm$6.0          & \textbf{75.7 $\pm$ 5.2} \\
    \midrule
    \midrule
    \rowcolor[HTML]{E6E6EF}
    \multicolumn{3}{c}{\textbf{ALL}}     & \textbf{56.9$\pm$1.7} & 23.5$\pm$1.5 & 12.4$\pm$1.1 & 55.4$\pm$1.7          & 51.5 $\pm$ 1.7 \\
    \bottomrule
  \end{tabularx}
  \vskip -0.2in
\end{table}

\subsection{Real Robot Experiments}
\label{subsec:real}

\looseness=-1
To validate the efficacy of our method in the real world, we use demonstrations of real wheeled robots for querying the prior dataset \DATASET{prior} constructed in simulated environments.
We then train the policy using a few real-robot data augmented with the retrieved simulation demonstrations.
We employ the custom-designed swarm robot platform inspired by Zooids~\cite{le2016zooids}, where the wheeled microrobots communicate with the host computer using $2.4$ GHz wireless communication.
The positions of all robots and an object are tracked in real time by a high-speed digital light processing (DLP) structured light projector system.
We can control the positions of the robots by sending 2D coordinates of their future locations from the host computer.

\begin{figure}[t]
    \begin{center}
    \includegraphics[width=0.85\linewidth]{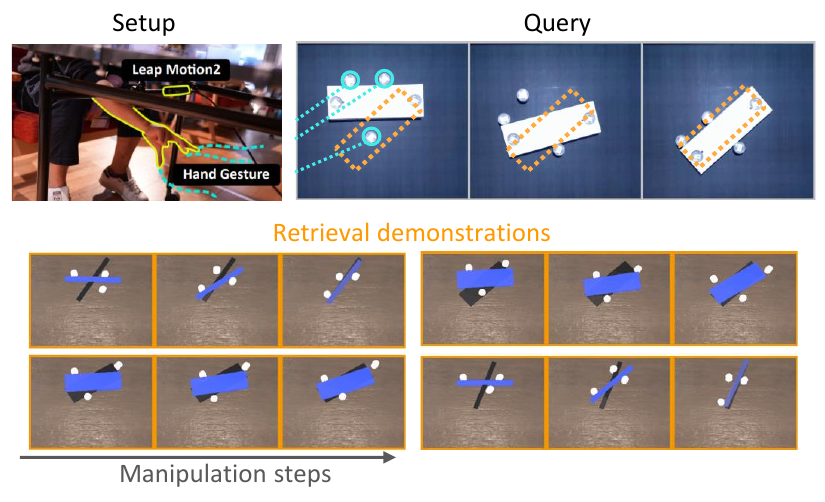}
    \vskip -0.0in
    \caption{In the above, we show the data collection setup of a hand gesture and a query that has been collected with hand gestures. In the below, we randomly sample four retrieval examples from 300 retrieved demonstrations.}
    \label{fig:retrieve_real}
    \end{center}
    \vskip -0.2in
\end{figure}

\begin{figure}[t]
    \begin{center}
    \includegraphics[width=0.9\linewidth]{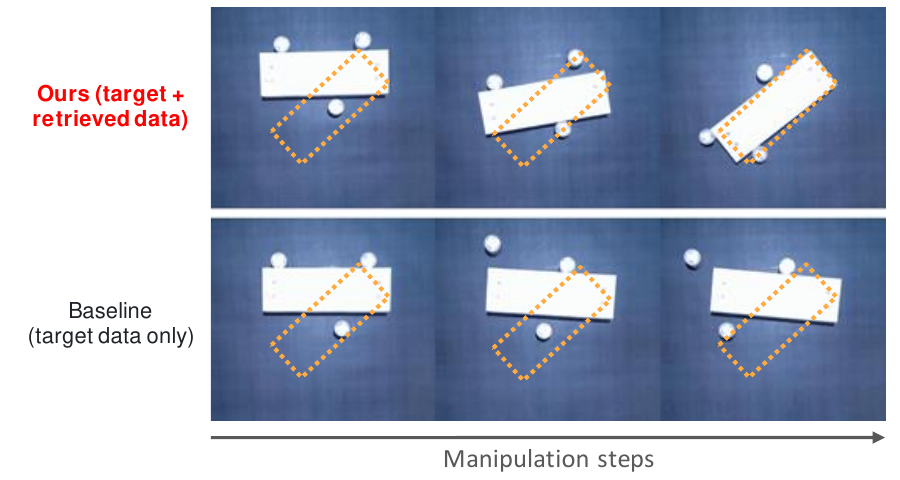}
    \vskip -0.0in
    \caption{Comparison highlighting the differences between our retrieve-and-learn policy and one that was trained using only real-world target trajectories.}
    \label{fig:real}
    \end{center}
    \vskip -0.2in
\end{figure}

We collect three demonstrations that navigate robots toward a given goal state by using a hand-gesture system with Leap Motion, where the hands' movement is translated into the trajectories of multiple robots.
Using the real-robot demonstrations as queries, we retrieve 300 simulated demonstrations from \DATASET{prior} for each query.
\Cref{fig:retrieve_real} shows four randomly sampled retrieval demonstrations in response to a query of the real-robot trajectory.
The result demonstrates that our method can accurately rank coordinated movements similar to the query higher.
\Cref{fig:real} compares our policy trained using the real and retrieved demonstrations and the one trained using only the real-robot demonstrations.
The results clearly demonstrate that the policy trained by our retrieve-and-learn framework successfully pushes an object closer to the goal state, while the policy trained using only a few real demonstrations fails to complete the task.

\section{Conclusion and Future Work}
\label{sec:conclusion}
We introduced a novel concept named Multi-Agent Coordination Skill Database, an essential skill representation repository for push manipulation distinctive to object shapes and its manipulating directions.
Extensive evaluations in the simulated environments clearly demonstrate the effectiveness of our retrieve-and-learn approach on the basis of the skill database both qualitatively and quantitatively.
There are two future directions. In real-world deployments, we have observed failure cases in scenarios involving complex collisions between multiple robots and a manipulation object. To address this issue, we believe that identifying the parameters through real-to-sim approaches or incorporating domain randomization techniques would be a promising direction. In addition, we aim to extend our framework's scope, examining its scalability for diverse push manipulation setups and its generalizability to other multi-agent tasks such as navigation and mobile-arm manipulation.

\bibliographystyle{./IEEEtran}
\bibliography{./IEEEabrv,./IEEE}

\end{document}